\documentclass[letterpaper, 10 pt, conference]{ieeeconf}  

\IEEEoverridecommandlockouts                         
\usepackage[table]{xcolor}
\usepackage{bm}
\usepackage{amssymb}
\usepackage{threeparttable}
\usepackage{wrapfig}
\usepackage{amsmath}
\usepackage{wrapfig}
\usepackage{lipsum}
\usepackage{siunitx}
\usepackage{booktabs}
\usepackage{glossaries}
\usepackage{hyperref}
\usepackage{graphicx}
\usepackage{url}
\usepackage{cleveref}
\usepackage{cite}
\usepackage{xcolor}
\usepackage{times} 

\title{\LARGE \bf
Probabilistic Differentiable Filters Enable Ubiquitous Robot Control with Smartwatches
}

\author{Fabian C Weigend, Xiao Liu, Heni Ben Amor
\thanks{All Authors are with the School of Computing and Augmented Intelligence, Arizona State University \texttt{$\lbrace$fweigend, xliu330, hbenamor$\rbrace$@asu.edu}%
}}

\begin{document}
\newacronym{gyro}{\ensuremath{\bm{\phi}}}{gyroscope measurements}
\newacronym{grav}{\ensuremath{\bm{\gamma}}}{gravity sensor}
\newacronym{lacc}{\ensuremath{\bm{\alpha}}}{linear acceleration sensor}
\newacronym{racc}{\ensuremath{\bm{\alpha}_\mathrm{raw}}}{raw acceleration}
\newacronym{swrot}{\ensuremath{\bm{\theta}}}{virtual rotation vector sensor}
\newacronym{swrot_calib}{\ensuremath{\bm{\theta}_c}}{calibration forward-facing direction}
\newacronym{pres}{\ensuremath{\rho}}{atmospheric pressure sensor}
\newacronym{rot_hip}{\ensuremath{\textbf{q}_\mathrm{h}}}{hip rotation}
\newacronym{rot_larm}{\ensuremath{\textbf{q}_\mathrm{l}}}{lower arm rotation}
\newacronym{rot_uarm}{\ensuremath{\textbf{q}_\mathrm{u}}}{upper arm rotation}
\newacronym{rot_larm_r}{\ensuremath{\textbf{q}^r_\mathrm{l}}}{relative lower arm rotation}
\newacronym{rot_uarm_r}{\ensuremath{\textbf{q}^r_\mathrm{u}}}{relative upper arm rotation}
\newacronym{6drr}{6DRR}{six-dimensional rotation representation}

\maketitle
\thispagestyle{empty}
\pagestyle{empty}

\begin{abstract}
Ubiquitous robot control and human-robot collaboration using smart devices poses a challenging problem primarily due to strict accuracy requirements and sparse information.
This paper presents a novel approach that incorporates a probabilistic differentiable filter, specifically the Differentiable Ensemble Kalman Filter (DEnKF), to facilitate robot control solely using Inertial Measurement Units (IMUs) from a smartwatch and a smartphone. The implemented system is cost-effective and achieves accurate estimation of the human pose state. Experiment results from human-robot handover tasks underscore that smart devices allow versatile and ubiquitous robot control. The code for this paper is available at \href{https://github.com/ir-lab/DEnKF}{\texttt{github.com/ir-lab/DEnKF}} and \href{https://github.com/wearable-motion-capture}{\texttt{github.com/wearable-motion-capture}}.
\end{abstract}

\section{Introduction}
Examining the human-robot relationship is a central concern in the field of artificial intelligence and robotics. 
To facilitate reliable human-robot collaboration, accurate estimations of the state of humans and robots are crucial. One challenge is that numerous robotics systems still rely heavily on costly motion capture systems in estimating the human pose, thereby restricting their suitability primarily to stationary setups or complex calibration procedures.

The gold standard for motion capture are cameras \cite{nagymate_application_2018}. These systems may feature multiple cameras on multiple base stations or in one device, e.g, Microsoft Kinect v2\cite{8462833} or Intel RealSense. 
These systems usually require a dedicated setup and suffer from line-of-sight issues. Alternatively, non-optical systems enable motion capture through Inertial Measurement Units (IMUs) \cite{DESMARAIS2021103275,DBLP:journals/corr/MarcardRBP17}. 
For this purpose, the Xsens motion capture system \cite{xsens} is commonly used \cite{yi2021transpose}. Also, Sony's recent Mocopi \cite{mocopi} promises new opportunities for non-visual real-time human motion capture. A downside of most IMU-based tracking is that they require users to carefully place multiple specialized units on their body and a calibration procedure.

Recent research investigates the opportunities of omnipresent IMUs in smartphones and smartwatches for motion capture. Some even limit themselves to a single device \cite{wei_real-time_2021,SmartwatchMiaomiao}. These approaches have promising potential for human-robot collaboration because motion capture through smart devices is ubiquitous and natural to the user \cite{weigend2023anytime}. However, leveraging the data of a single smartwatch for motion capture of sufficient accuracy for robot control is challenging, and previous work had to constrain the user to a constant body forward-facing direction \cite{weigend2023anytime}. 

This work builds upon \cite{weigend2023anytime} and incorporates the sensor data of a connected smartphone to enable ubiquitous robot control without body orientation constraints. Recursive Bayesian filters, particularly Kalman filters, play a pivotal role in tasks such as predicting the future movements of human interaction partners~\cite{WANG2022102310}, tracking objects over time~\cite{chen2011kalman}, and ensuring stability during robot locomotion~\cite{reher2019dynamic}. We propose that advances in state estimation, specifically the Differentiable Ensemble Kalman Filter (DEnKF) \cite{liu2023enhancing}, allow to enhance the accuracy and stability of motion capture using ubiquitous smart devices. 

\begin{figure}[t!]
\centering
\includegraphics[width=\linewidth]{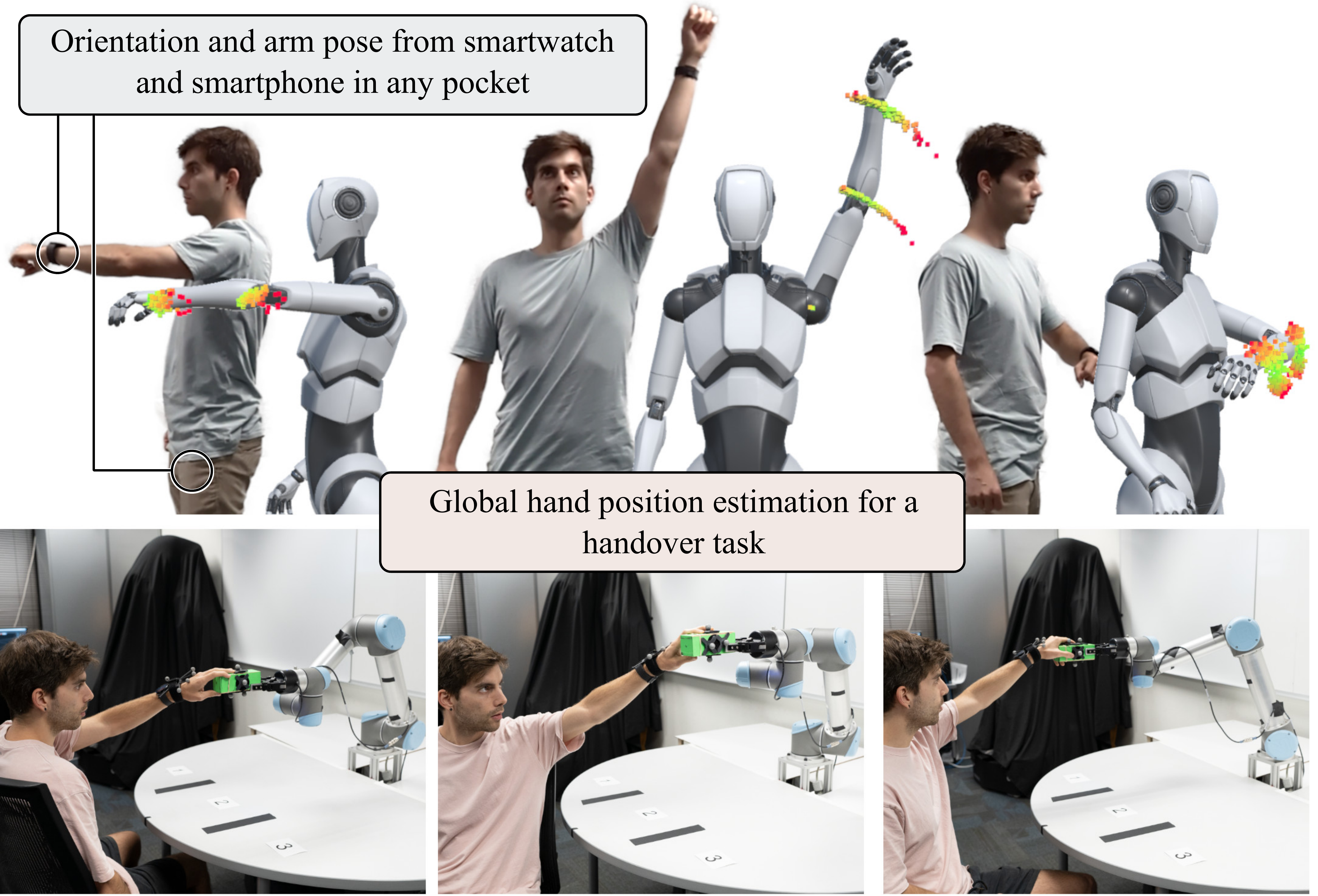}
\caption{{\bf Top}: Our Differentiable Ensemble Kalman Filter (DEnKF) achieves robust body orientation and arm pose estimations from the sensor data of a single smartwatch and a smartphone. The user can place the phone in any pocket. {\bf Bottom}: Orientation predictions are also accurate when the user sits. We evaluate pose predictions on test data and in a human-robot collaboration handover task.}
\label{fig:overview}
\end{figure}

As depicted in \Cref{fig:overview}, using a probabilistic differentiable filter provides us with a distribution of solutions, aiding stability and adding an important measure of uncertainty for robot control. The proposed filter allows to strike the balance between less-constrained movements while still achieving stable and effective pose estimations suitable for human-robot collaboration. 





\section{Methodology}

We present our approach by describing our data collection and defining our states and observations in \Cref{subsec:col_obs_state}. Then, we define the used Differentiable Ensemble Kalman Filter (DEKnF) for human pose state estimations in \Cref{subsec:diff_kal}.
 


\subsection{Data Collection, Observation and State}
\label{subsec:col_obs_state}

For data collection, we develop two apps and make use of a research-grade motion capture system. The apps for the smartwatch and smartphone stream sensor data to a remote machine.

\textbf{Observation:} We define that the raw observation $\bf{y}$ consists of the following values 
${\bf y} = [ \Delta t, \pmb{\theta}_{\mathrm{sw}}, 
    \pmb{v}, 
    \pmb{\alpha},
    \pmb{\gamma},
    \pmb{\phi},
    \rho,
    \mathbf{r}_h
    ]^\top$, 
with $\bf{y} \in \mathbb{R}^{22}$, where $\pmb{\alpha},\pmb{\gamma},\pmb{\phi} \in \mathbb{R}^3$ are the IMU readings. Namely, these are the average values of the linear acceleration, gravity, and gyroscope since the previous observation $\Delta t$ seconds ago. The averaging is necessary because the apps stream data at around 80\,Hz and the accelerometer and gyroscope record measurements at a faster rate. We also integrate linear acceleration measurements between two observations and denote them as velocities $\pmb{v}$. In addition, the apps record the \mbox{\gls{swrot}}, which is provided by Android and Wear OS. We record this orientation in form of a continuous \gls{6drr}, which is well-suited for training neural networks~\cite{zhou_continuity_2019}. 

\textbf{Calibration:} When the user starts the app on their smartwatch, they usually hold their arm up, parallel to the chest and hip. We use this start position to calibrate \gls{swrot} with the first initial orientations $\gls{swrot}_\text{init}$, which gives us the calibrated orientation $\gls{swrot}_\text{sw}$. The value $\rho$ with $\rho \in \mathbb{R}$ is the atmospheric pressure sensor. We also calibrate this one by subtracting the reading when the user started the app from all subsequent ones. Finally, the phone provides the body-forward facing direction. Also here, we record the orientation sensor of the phone and calibrate it with the first observation in the starting position. Further, we assume that the user has their forward-facing direction parallel to the watch at the starting position. This allows us to estimate an offset rotation from the phone to the forward direction, bringing it into the same global reference frame as the watch. From this, we denote the body orientation as the sine and cosine of the calibrated up-axis rotation of the phone, represented as $\bf{r}_h\in\mathbb{R}^{2}$ in our raw observation $\bf{y}$.

\begin{figure}[t!]
\centering
\includegraphics[width=\linewidth]{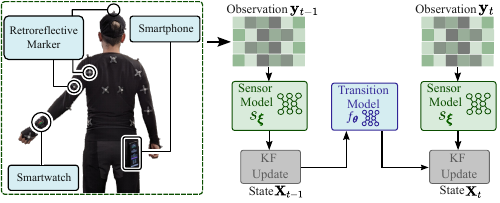}
\caption{{\bf Left}: Our data collection setup with smart devices and Optitrack system. {\bf Right}: The DEnKF model structure. The sensor model projects raw observations to the observation space, the stochastic transition model forwards the ensemble one step in $t$, and the KF update step corrects the state.}
\label{fig:method}
\vspace{-0.2in}
\end{figure}

\textbf{State:} Our state entails the arm pose and body forward-facing direction of the human. The ground truth values were recorded with the research-grade optical motion capture system OptiTrack \cite{nagymate_application_2018}. We record data from participants who wore a 25-marker-upper-body suit along with the smartwatch on their left wrist and a smartphone in their pocket. We collect the \mbox{\gls{rot_uarm}}, \mbox{\gls{rot_larm}} also in the continuous \gls{6drr} and the body-forward facing direction $\bf{r}_h$. Therefore, the ground truth state for motion capture is denoted as $\mathbf{x} = [\text{\gls{rot_larm}}, \text{\gls{rot_uarm}}, \mathbf{r}_h]$, where $\mathbf{x} \in \mathbb{R}^{14}$. In our Bayesian filtering framework, we define the learned observation $\tilde{\mathbf{y}}$ to be equivalent to $\mathbf{x}$.

\subsection{Differentiable Ensemble Kalman Filter}
\label{subsec:diff_kal}
To model the temporal transition of the human pose $\mathbf{x}$ and map raw observations $\mathbf{y}$ to the state space, we utilize DEnKF~\cite{liu2023enhancing,liu2023learning,liu2023alphamdf} as shown in \Cref{fig:method}. This state estimation approach enables us to learn and infer the dynamics of the human pose over time, while efficiently incorporating and processing the observed data. We maintain the core algorithmic steps of an Ensemble Kalman Filter (EnKF)~\cite{evensen2003ensemble} while leveraging the capabilities of stochastic neural networks (SNNs)~\cite{gal2016dropout}. There are two steps in DEnKF, the \emph{Prediction Step} propagates the state one step further in time, and the \emph{Update Step} corrects the state based on newly collected observations.
Let ${\bf X}_{0:N}$ denote the states of $N$ steps in $t$ with number of $E$ ensemble members, we initialize the filtering process with ${\bf X}_{0:N} = [ {\bf x}^{1}_{0:N}, \dots, {\bf x}^{E}_{0:N}]$, where $E \in \mathbb{Z}^+$. 

\textbf{Prediction Step}: In this step, the Transition Model takes the previous states and predicts the next state. We use a window of $N$ and we leverage the stochastic forward passes from a trained state transition model to update each ensemble member: 
    \begin{equation}
    \begin{aligned}\label{eq:1}
          {\bf x}^{i}_{t} & \thicksim  f_{\pmb {\theta}} ({\bf x}^{i}_{t}|{\bf x}^{i}_{t-N:t-1}),\  \forall i \in E.
    \end{aligned}
   \end{equation}
 Matrix ${\bf X}_{t} = [{\bf x}^{1}_{t}, \cdots, {\bf x}^{E}_{t}]$ holds the updated ensemble members which are propagated one step forward through the state space. Note that sampling from the transition model $f_{\pmb {\theta}}(\cdot)$ implicitly introduces a process noise.

\textbf{Update Step}: Given the updated ensemble members ${\bf X}_{t}$, a nonlinear observation model $h_{\pmb {\psi}}(\cdot)$ is applied to transform the ensemble members from the state space to observation space. The observation model is realized via a neural network with weights $\pmb {\psi}$:
    \begin{align}
    \label{eq:2}
        {\bf H}_t {\bf X}_{t} &= \left[ h_{\pmb {\psi}}({\bf x}^1_{t}), \cdots, h_{\pmb {\psi}}({\bf x}^E_{t}) \right],\\
        \label{eq:3}
        {\bf H}_t {\bf A}_{t} &=  {\bf H}_t {\bf X}_{t} 
        - \left[\frac{1}{E} \sum_{i=1}^E h_{\pmb {\psi}}({\bf x}^i_{t}),
        \cdots,
        \frac{1}{E} \sum_{i=1}^E h_{\pmb {\psi}}({\bf x}^i_{t})\right]. \nonumber
    \end{align}
${\bf H}_t {\bf X}_{t}$ is the predicted observation, and ${\bf H}_t {\bf A}_{t}$ is the sample mean of the predicted observation at $t$. EnKF treats observations as random variables. Hence, the ensemble can incorporate a measurement perturbed by a small stochastic noise thereby accurately reflecting the error covariance of the best state estimate~\cite{evensen2003ensemble}. As shown in \Cref{fig:method}, we incorporate a Sensor Model that can learn projections between the learned observation and raw observation space. To this end, we leverage the methodology of SNN to train a stochastic sensor model that takes N steps of the raw observation and predicts the current learned observation using $s_{\pmb {\xi}}(\cdot)$:
    \begin{equation}
    \begin{aligned}\label{eq:sensor}
          \tilde{{\bf y}}^{i}_t & \thicksim  s_{\pmb {\xi}} (\tilde{{\bf y}}^{i}_t|{\bf y}_{t}),\  \forall i \in E,\\
    \end{aligned}
   \end{equation}
where ${\bf y}_{t}$ represents the noisy observation. Sampling yields observations $\tilde{{\bf Y}}_t = [\tilde{{\bf y}}^{1}_t, \cdots, \tilde{{\bf y}}^{E}_t]$ and sample mean \mbox{$\tilde{{\bf y}}_t = \frac{1}{E}\sum_{i=1}^i\tilde{{\bf y}}^i_t$}.  The innovation covariance ${\bf S}_t$ can then be calculated as:
    \begin{equation}
    \begin{aligned}\label{eq:4}
        {\bf S}_t &= \frac{1}{E-1}  ({\bf H}_t {\bf A}_t)  ({\bf H}_t {\bf A}_t)^T + r_{\pmb {\zeta}}(\tilde{{\bf y}_t}),
    \end{aligned}
    \end{equation}
where $r_{\pmb {\zeta}}(\cdot)$ is the measurement noise model implemented using MLP. We use the same way to model the observation noise as in~\cite{kloss2021train}, $r_{\pmb {\zeta}}(\cdot)$ takes an learned observation $\tilde{{\bf y}_t}$ in time $t$ and provides stochastic noise in the observation space by constructing the diagonal of the noise covariance matrix. The final estimate of the ensemble ${\bf X}_{t|t}$ can be obtained by performing the measurement update step:
    \begin{equation}
    \begin{aligned}\label{eq:5}
        {\bf A}_t = {\bf X}_{t} - \frac{1}{E}\sum_{i=1}^E&{\bf x}^i_{t},\ {\bf K}_t = \frac{1}{E-1} {\bf A}_t ({\bf H}_t {\bf A}_t)^T {\bf S}_t^{-1},\\
        {\bf X}_{t|t} &= {\bf X}_{t} + {\bf K}_t (\tilde{{\bf Y}}_t - {\bf H}_t {\bf X}_{t}),
    \end{aligned}
    \end{equation}
where ${\bf K}_t$ is the Kalman gain. In inference, the ensemble mean ${\bf \bar{x}}_{t|t} = \frac{1}{E}\sum_{i=1}^E {\bf x}^i_{t|t}$ is used as the updated state. 

\textbf{Prediction Targets:} Once the estimated state, comprising the rotation values for both the lower arm and upper arm, is obtained, we utilize forward kinematics with fixed lower arm length $l_l$ and fixed upper arm length $l_u$ to determine the corresponding Cartesian XYZ coordinates of the wrist. This is done due to the constraint of a fixed sagittal plane orientation $\mathbf{r}_h$, where the elbow position is limited to a sphere around the shoulder with a radius of $l_u$. Additionally, the wrist position must lie on a manifold defined by spheres with a radius of $l_l$ around all possible elbow positions, as described in prior work~\cite{shen_i_2016}.

\textbf{Training:} DEnKF contains four sub-modules: a state transition model, an observation model, an observation noise model, and a sensor model. The entire framework is trained in an end-to-end manner via a mean squared error (MSE) loss between the ground truth state $\hat{{\bf x}}_{t|t}$ and the estimated state ${\bf \bar{x}}_{t|t}$ at every timestep. We also supervise the intermediate modules via loss gradients $\mathcal{L}_{f_{\pmb {\theta}}}$ and $\mathcal{L}_{s_{\pmb {\xi}}}$. Given ground truth at time $t$, we apply the MSE loss gradient calculated between $\hat{{\bf x}}_{t|t}$ and the output of the state transition model to $f_{\pmb {\theta}}$ as in Eq.~\ref{eq:loss1}. We apply the intermediate loss gradients computed based on the ground truth observation $\hat{{\bf y}_t}$ and the output of the stochastic sensor model $\tilde{{\bf y}}_t$: 
    \begin{align}
    \label{eq:loss1}
    \mathcal{L}_{f_{\pmb {\theta}}} =  \| {\bf \bar{x}}_{t|t-N:t-1} - \hat{{\bf x}}_{t|t}\|_2^2,\ \ 
        \mathcal{L}_{s_{\pmb {\xi}}} =\| \tilde{{\bf y}_t} -  \hat{{\bf y}_t}\|_2^2.
    \end{align}
All models in the experiments were trained for 50 epochs with batch size 256, and a learning rate of $\eta = 10^{-5}$. We chose the model with the best performance on a validation set for testing. The ensemble size of the DEnKF was set to 32 ensemble members.

\section{Evaluation}

We discuss the validation accuracy on a separate test dataset. Secondly, we assess the performance of our model by applying it in a human-robot handover tasks with a real UR5 robot.

\subsection{Evaluation on Dataset}
\begin{figure}[t]
\centerline{\includegraphics[width=0.5\textwidth]{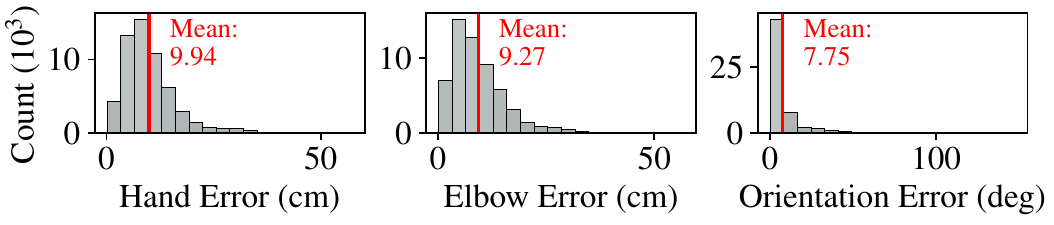}}
\caption{Prediction error distributions of our DEnKF on the test dataset.}
\label{fig:test_data_errors}
\vspace{-0.1in}
\end{figure}

The training dataset comprises data collected from five human subjects. Written informed consent was obtained and approved by the institutional review board~(IRB) of ASU under the ID~STUDY00017558. Each subject was instructed to wear the smartwatch on their left arm, the smartphone in one of their pockets and a 25-marker Optitrack suit while performing free random arm motions. The subjects were also encouraged to change their body forward-facing direction and to move around in the area covered by the optical tracking system. In total, we gathered a dataset of 970,493 data points. To further augment the dataset, we artificially adjusted the calibrated smartwatch and smartphone data to simulate new body orientations. This is possible because remaining sensor measurements, e.g., accelerometer or barometer, are in the reference frame of the watch. In total, our augmented training dataset amounts to 4,259,746 data points.

\begin{figure*}[t!]
\centering
\includegraphics[width=\linewidth]{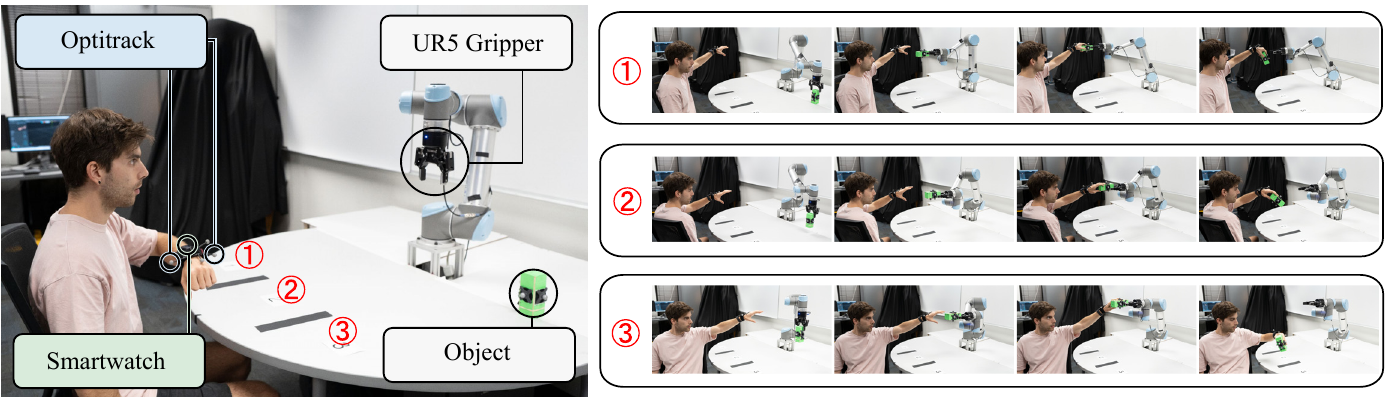}
\caption{The handover task system setup is illustrated on the {\bf left}, showcasing the workspace divided into three distinct labeled areas. On the {\bf right}, picture sequences of the handover results are presented for each of these designated zones.}
\vspace{-1\baselineskip}
\label{fig:handover}
\end{figure*}

We evaluate the DEnKF prediction accuracy on a test dataset completely separate from the training process. The test dataset comprises 26,688 observations. The participants were asked to perform a series of movements, including arm swing, arm cross on chest and behind the head, arm raise, waving, boxing, clapping, walking in a figure-eight pattern, and jogging in a circular path. \Cref{fig:test_data_errors} summarizes the performance of the DEnKF model on the test dataset. On average, hand positions are off by 9.94\,cm, elbow positions by 9.27\,cm, and body orientation by 7.75\,deg. The DEnKF quantifies the state uncertainty through the distribution of ensemble members, for example, depicted at the top in \Cref{fig:overview}. Our framework achieves inference speed at $\sim$62\,Hz on a system using an Intel® Xeon(R) W-2125 CPU and NVIDIA GeForce RTX 2080 Ti. 

\begin{table}[ht]
\begin{center}
\footnotesize
\caption{Model Comparison with related work}
\label{Tab:model_compare}
\scalebox{0.95}{
\begin{tabular}{c c c S[table-format=2.2] c c}
\toprule 
 &  & Free Forward-& \multicolumn{1}{c}{Wrist} & Elbow & Hip\\
Method & Anywhere & Facing Dir. & \multicolumn{1}{c}{(cm)} & (cm) & (deg) \\
\midrule
\cite{wei2021real} & $\times$ & $\checkmark$ & 10.93 & - & -\\
\cite{liu2022real} & $\checkmark$ & $\times$ & 8.50 & 8.50 & -\\
\cite{shen_i_2016} & $\checkmark$ & $\times$ & 9.20 & 7.90 & -\\
\cellcolor{gray!10} Ours & \cellcolor{gray!10}$\checkmark$ & \cellcolor{gray!10}$\checkmark$ & \cellcolor{gray!10}9.94 & \cellcolor{gray!10}9.27 & \cellcolor{gray!10}7.75\\
\bottomrule
\end{tabular}}
\end{center}
\vspace{-0.1in}
\end{table}

We compare our results with other related works~\cite{wei2021real,liu2022real,shen_i_2016} in \Cref{Tab:model_compare}. 
The method of \cite{wei2021real} requires inference in the same environment where the training data was collected, therefore, it is not applicable \emph{anywhere}. Further, \cite{wei2021real} focuses on pose predictions for the wrist, omitting the rest of the body pose including elbow or hip orientation. Methods of \cite{liu2022real} and \cite{shen_i_2016} demonstrate lower errors for wrist and elbow but fix the user to a constant forward-facing direction. In contrast, our proposed method using the DEnKF also provides an estimate of the Hip pose and allows for ubiquitous pose estimation regardless of location or changes in body orientation.

\subsection{Handover Task}
We demonstrate the efficacy of the trained model in a human-robot handover task. As depicted in \Cref{fig:handover}, participants sit in a chair and engage in handover interactions with the robot. Participants are free to rotate on the chair to experiment with various handover scenarios. The smartphone in their pocket and the smartwatch on their wrist enable estimating the body orientation and arm pose to extract the global hand position.

{\bf Task Setup}: Participants perform six handover interactions. Each handover interaction is treated as an individual task. At the onset of each task, the participant holds their hand as shown on the left in \Cref{fig:handover} and initiates the task by issuing a voice command. Like in the work of \cite{weigend2023anytime}, the smartwatch recognizes the voice command and triggers the robot. Subsequently, the robot grasps the green cube and moves it towards the tracked hand position of the participant. When the cube reaches close proximity to the hand, the participant says ``give me the cube" and closes their hand around it. In response to this voice command, the robot releases its grip on the cube, signifying the successful completion of the handover.

To evaluate a range of handover positions, we divide the table surface into three distinct areas labeled as 1, 2, and 3. Each participant is instructed to perform two handovers in each area, with one at a high position and one at a low position. In total, every participant performs all six handovers utilizing smartwatch and smartphone tracking, along with an additional six handovers conducted with the gold-standard Optitrack for our baseline comparison. The order of the tasks and tracking modes is randomized to eliminate any potential biases. 

{\bf Results}: The handover experiment utilizing smartwatches is depicted in \Cref{fig:handover} (right), illustrating the handover process conducted in zones 1, 2, and 3. We collected data from five human subjects, with each subject performing a total of 12 handover tasks. 
\begin{table}[t]
\begin{center}
\footnotesize
\caption{Handover task results}
\label{Tab:handover}
\begin{tabular}{ 
c
S[table-format=2.2] @{${}\pm{}$} S[table-format=1.2]
S[table-format=1.2] @{${}\pm{}$} S[table-format=1.2]
}
\toprule 
Method & \multicolumn{2}{c}{Time (s)} & \multicolumn{2}{c}{Dist. (cm)}\\
\midrule
Wearable &11.74 & 5.35 & 7.74 & 4.68 \\
Optitrack &9.86 & 2.67 & 7.77 & 4.71 \\
Difference &1.88 & 4.83 & 0.03 & 7.13 \\
\bottomrule
\end{tabular}
\end{center}
\end{table}
The evaluation metrics in \Cref{Tab:handover} include the task completion time and the actual 3D distance between hand and cube when the participant gave the voice command to hand over the cube.
As summarized in \Cref{Tab:handover}, we observe a relatively small disparity in the task completion time, with an average difference of 1.88 seconds. Further, observed handover 3D distances exhibit a minimal average difference of 0.03\,cm when comparing both methods. Altogether, the results suggest that handover tasks with smartwatch and smartphone tracking might take about 1.88 seconds longer but the participant is comfortable to complete the handover at similar distances. It is important to note that all handover tasks were accomplished successfully, with users consistently grasping the cube within the desired zones without any instances of dropping it. The obtained results indicate that the smartwatch system is comparable to the Optitrack system for this task, thus establishing its potential as a cost-efficient alternative.

\section{Conclusion}
This work introduces the integration of a ubiquitous robot control system by leveraging smart devices and employing differentiable filters. The experimental results demonstrate that the proposed framework effectively addresses the estimation challenges related to human arm pose estimation, especially in scenarios involving diverse human hip orientations. Additionally, the results from the human-robot handover task showcase that the proposed system achieves comparable error metrics, highlighting its effectiveness. With no additional user instrumentation, the proposed framework offers new and intriguing possibilities for low-cost robot control and human-robot collaboration applications.

\bibliographystyle{IEEEtran}
\scriptsize{
\bibliography{references}
}

\end{document}